\documentclass{article} 
\usepackage{iclr2019_conference,times}

\usepackage{amsmath,amsfonts,bm}

\def\eqref#1{equation~\ref{#1}}

\def\1{\bm{1}}

\DeclareMathAlphabet{\mathsfit}{\encodingdefault}{\sfdefault}{m}{sl}
\SetMathAlphabet{\mathsfit}{bold}{\encodingdefault}{\sfdefault}{bx}{n}

\usepackage{natbib}
\usepackage{url}
\usepackage{graphicx}
\usepackage{hhline}
\usepackage{multirow}
\usepackage{wrapfig}

\title{Salient Skin Lesion Segmentation via Dilated Scale-Wise Feature Fusion Network} 

 \author{Pourya Shamsolmoali \thanks{pshams55@gmail.com} ~ ,  Masoumeh Zareapoor     \\
Shanghai Jiao Tong University     \\
 Shanghai, China        \\
\AND
Eric~Granger   \hspace{7.7cm} Huiyu Zhou \\
\'Ecole de technologie sup\'erieure, Universit\'e du Qu\'ebec  \hspace{2cm} University of Leicester\\
 Montreal, Canada  \hspace{7.25cm} Leicester, UK \\
}

\iclrfinalcopy 
\begin{document}

\maketitle

\begin{abstract}
Skin lesion detection in dermoscopic images is essential in the accurate and early diagnosis of skin cancer by a computerized apparatus. Current skin lesion segmentation approaches show poor performance in challenging circumstances such as indistinct lesion boundaries, low contrast between the lesion and the surrounding area, or heterogeneous background that causes over/under segmentation of the skin lesion. To accurately recognize the lesion from the neighboring regions, we propose a dilated scale-wise feature fusion network based on convolution factorization. Our network is designed to simultaneously extract features at different scales which are systematically fused for better detection. The proposed model has satisfactory accuracy and efficiency. Various experiments for lesion segmentation are performed along with comparisons with the state-of-the-art models. Our proposed model consistently showcases state-of-the-art results. 
\end{abstract}

\section{Introduction}
Skin melanoma cases represent about 1\% of skin cancers but cause most of the skin cancer deaths. According to available data, 192,310 new melanoma cases, 95,830 non-invasive and 96,480 invasive, were reported in the United States between 2019-2020, resulting in around 10,000 deaths\footnote{https://www.aad.org/media/stats-skin-cancer}. To automatically detect melanoma in dermoscopy images, it is imperative to precisely outline the boundary of the lesion in order to carry out proper analysis of the clinical indicators presented in the lesion \citep{garcia2019segmentation}. There are various challenges with effective skin lesion detection in dermoscopy images including variations of lesion size, unclear boundaries, lesion types and different skin colors. Complex background regions (for instance, heterogeneous background \citep{barata2018survey}) in dermoscopic images also make it challenging to recognize the lesions and distinguish them from neighbouring areas. For example, specular reflections and hair are other barriers that reduce the accuracy of the segmentation.
Indeed, significant efforts have been carried out in recent years and achieved good progress in this field \citep{barata2018survey}. In \citep{celebi2015state}, the authors presented a comprehensive review on detection of lesion border in dermoscopy images.\\
 
Salient Object Detection (SOD), as a useful method, can pinpoint the location and outline of the most salient regions or objects in an image. SOD can significantly increase the performance of different computer vision tasks, like object detection~\citep{song2018pyramid, wang2019salient} and segmentation~\citep{mehta2018espnet}. 
Currently, convolutional neural networks (CNNs) models have emerged as a promising support for SOD and regularly reported major improvements~\citep{song2018pyramid, wang2019salient, mehta2018espnet}. Despite this progress, how to devise an efficient yet effective CNN model for SOD remains an open challenge. 
\begin{figure}
\centering
\includegraphics[width=11cm]{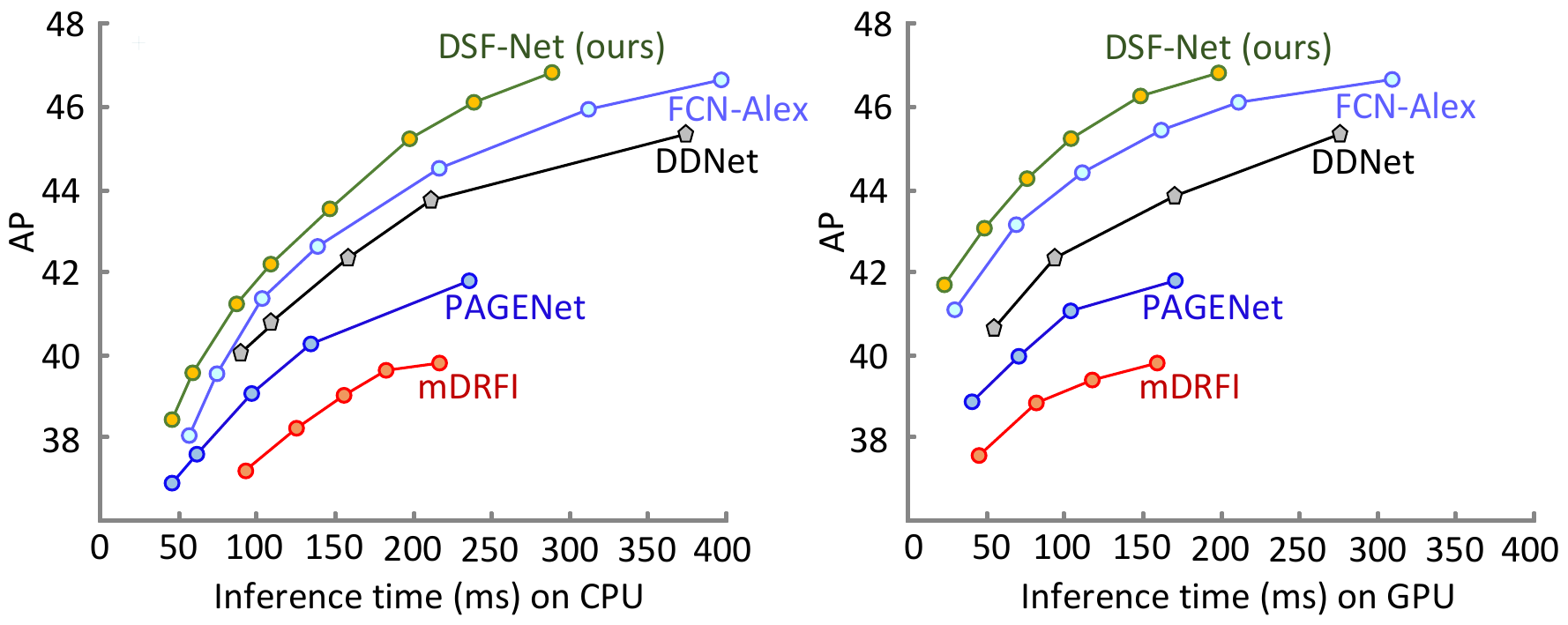}
\caption{Average Precision (AP) vs. inference time per image. The green curve shows results of DSF-Net which is described in the text body.}
\label{fig:1}
\end{figure}
To address the above problem, we propose a {\em dilated scale-wise feature fusion (DSF)} for skin lesion segmentation, to significantly reduce the computational complexity of CNNs~\citep{mehta2018espnet, xie2017aggregated, szegedy2017inception}. Based on the DSF units, we introduce a network called, {\tt DSF-Net}, that can be used for semantic segmentation. DSF-Net has achieved satisfactory segmentation accuracy with minor latency. 

\begin{figure*}
\centering
\includegraphics[width=13 cm]{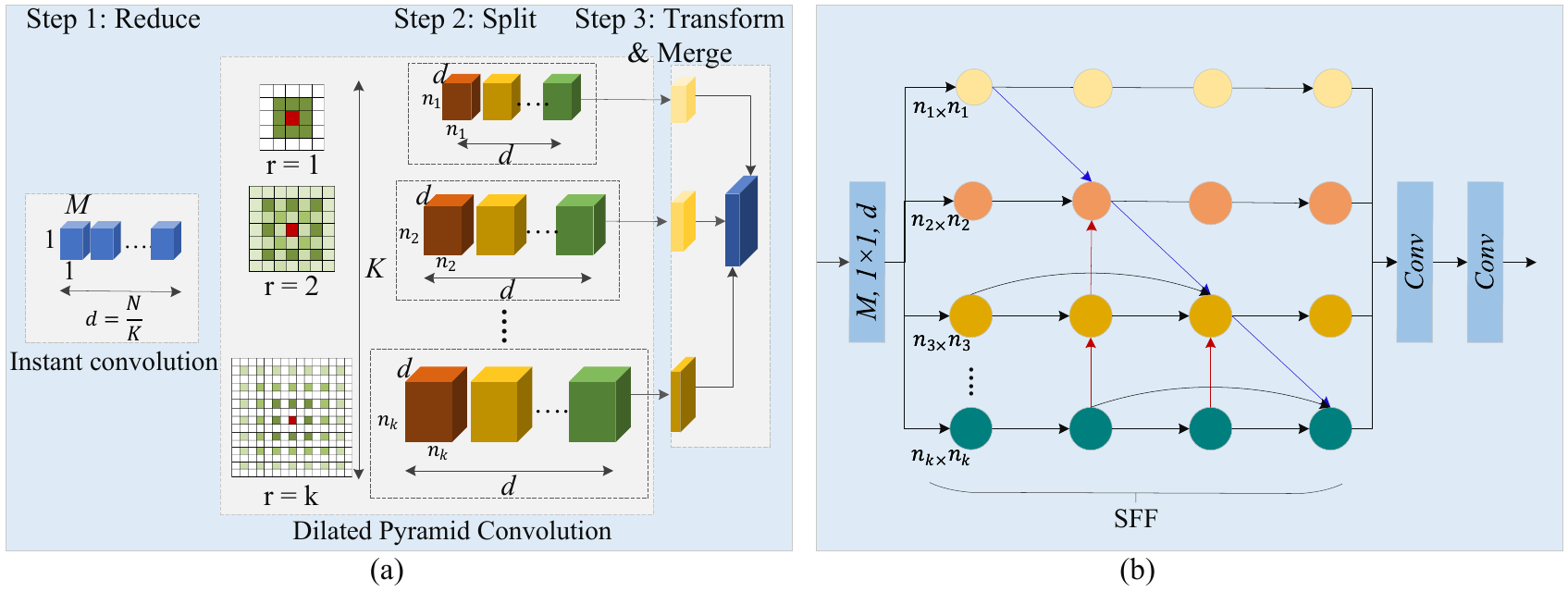}
\caption{Diagram of the proposed DSF model: (a) the DSF unit is decomposed into instant convolution and dilated pyramid convolutions; (b) Block diagram of pyramid unit.}
\label{fig:2}
\end{figure*}

Fig.~\ref{fig:1} shows a summary of comparison results.  DSF decomposes a typical convolution operation into two stages:  instant convolutions and dilated pyramid convolutions, as shown in Fig.~\ref{fig:2}. To reduce the memory usage of our proposed model while training, we do not use data augmentation. However, to increase the robustness, we use an instant convolution, where the weight matrix is initialized as a random rotation matrix \citep{dohmatob2022non}. This simple method improves the generalization and robustness to domain distribution. The instant convolutions are also effective in reducing the computational cost, whereas the dilated pyramid networks re-sample the feature maps in order to widely learn from a large and more useful receptive field. 
In addition, scale-wise feature fusion is adopted to aggregate different scale features. We show that the proposed DSF model has better segmentation accuracy and efficiency compared to other factorized convolution models. It has been observed that the existing pyramid models~\citep{song2018pyramid, wang2019salient, mehta2018espnet} have high computational costs and cannot be applied at different CNN layers and at any spatial level for representation learning. In contrast to the above mentioned methods, DSF unit has lower computation costs, therefore, can be efficiently applied at any spatial layer of a CNN. Current dilated convolution models~\citep{song2018pyramid, shamsolmoali2019image, shamsolmoali2019single} are inefficient, but our proposed DSF model utilizes an efficient dilated convolutions. To evaluate the performance of DSF-Net, comprehensive experiments on four publicly accessible datasets have been conducted. DSF-Net demonstrates to be more accurate and efficient than the other state-of-the-art.

\section{Related Work}
Accurate segmentation of the lesion from the normal skin is useful for effective skin cancer diagnosis~\citep{unver2019skin}. The segmentation methods are categorised to three groups. Histogram thresholding models apply a threshold value for the lesion segmentation~\citep{garcia2019segmentation, unnikrishnan2007toward}. Unsupervised models, unlike the supervised approaches, do not need any training. They utilize different image processing approaches to enhance the colour, and spatial location which have a significant impact on visual attention~\citep{ahn2017saliency}. The last one is the supervised segmentation models. Such approaches segment the skin lesions by training a detection network such as SVM and artificial neural networks~\citep{xie2013automatic}. But, these traditional methods cannot achieve satisfactory segmentation results and unable to overcome the problems like unclear lesion boundaries, and hair. Recently, CNNs have achieved a great progress in detection, and segmentation of objects~\citep{wang2019salient, mehta2018espnet, xie2017aggregated}.\\

In ~\citep{kaymak2020skin}, a model based fully convolution network (FCN) is proposed for skin lesion segmentation. In the implementation, a new FCN architecture is designed based on the AlexNet ~\citep{krizhevsky2012imagenet} called FCN-AlexNet. A variational autoencoder is used to label skin tone of lesion images \citep{bevan2022detecting} and this framework is used to annotate the benchmark datasets for lesion detection. This is a useful method for data annotation, nevertheless, this model does not perform effectively for complex images.
In ~\citep{ liu2021skin} a CNN architecture based on auxiliary information is proposed, where edge prediction is conducted as an auxiliary task alongside the segmentation task. Furthermore, a multi-scale feature aggregation module is suggested, which utilizes features of several scales. These models, however, perceive dark regions to be regions of interest, and their segmentation performance is low when the region of interest is brighter than the surrounding skin region. ~\citep{ sarker2021slsnet} proposes a lightweight generative adversarial network that integrates 1-D kernel weighted networks, position, and channel attention. The position and channel attention modules, respectively, improve the discriminative ability between lesion and non-lesion feature representations in spatial and channel dimensions. In ~\citep{akyel2022linknet, kosgiker2022significant}, CNNs based methods are introduced for noise reduction and lesion segmentation. Noise reduction such as hair removal is important for a correct segmentation of the lesions. However, these models require complex parameter adjustment that increases the computation cost. 
The key reason for the success of CNNs is due to their ability in extracting deep and robust features from the input images and their hierarchical feature learning. Several approaches such as network compression and factorized convolutional kernels have been introduced to accelerate CNNs ~\citep{mehta2018espnet, szegedy2017inception}.\\ 

\noindent {\bf Convolution factorization } is an effective approach which breaks down the convolutional operation into multiple stages to ensure a fast computation. Factorization has demonstrated its ability in dropping the computational complexity for deep CNNs such as Inception~\citep{szegedy2017inception}, and factorized network~\citep{jin2014flattened}. \\

\noindent {\bf Dilated convolution }~\citep{shamsolmoali2019image,shamsolmoali2019single} is a type of conventional convolution in which the effective receptive field of the kernel is enlarged by adding zeros between each pixel in the convolutional kernel. As an example, for $n\times n$ dilated convolution kernel, the operative size of the kernel is $[(n-1) r+1]^2$.

\section{Proposed Dilated Scale-Wise Feature Fusion Network}
As shown in Fig.~\ref{fig:2}(a), the proposed {\tt DSF-Net} is composed of an instant convolution and dilated pyramid convolutions. A a $1\times1$ instant convolution is used to increase the robustness and reduce the dimension of feature maps. Then, the dilated pyramid convolutions resample the rescaled feature maps with $n\times n$ dilated convolutional $K$ kernels, in which the dilation rate is $2^{(k-1)}$ where $k={1,2,...,k}$. In our model the factorization technique considerably decreases the memory usage and the number of parameters, while keeping the potential receptive fields $[(n-1)^{2^{k-1}}+1]^2$ . In the proposed DSF unit, each kernel of the dilated convolution uses multiple receptive fields for weights learning. A typical convolution layer receives an input $F_i \in R^{H\times W\times M}$ and has $N$ kernels $K\in R^{m\times n\times M}$ for generating a feature map $F_0 \in R^{(H\times W\times N)}$, in which {\it H, W} indicate the feature map dimension, while {\it m , n} are the dimension of the kernel (width and height), and {\it M , N} are the total input and output feature channels. For better understanding, we assume $m = n$. The ordinary convolution kernels learn ($n^2 MN$) parameters. Indeed, these parameters are based on the kernels dimension, the number of input $M$ and output $N$ channels.\\

{\bf Width splitting and Network Design}: To increase the efficiency of the proposed DSF unit, the hyperparameter $K$ is introduced. $K$ participates to split the dimension of the feature maps consistently through each DSF unit. For a certain $K$, the DSF unit in the first step decreases the feature maps from $M$ dimensions to $\frac{N}{K}$ dimensions as shown in Fig.~\ref{fig:2}{(a)}. Then, the feature maps shrink between $K$ branches that have different scales. In each branch of the saliency network, a pyramid attention module is placed which is integrated to generate high discriminative feature maps. Different from current saliency models that consider all the points equally in a saliency feature map, DSF only focuses on the important regions features and reuses the multi-scale information by several attention layers which take advantage of scale-wise features to build a uniform pyramid network. To be more precise, let $X$ represent a feature map extracted from a convolution layer of a saliency network with $C$ channels, $X\in {R^{M\times N\times C}}$. We aim to learn a group of same size attention masks that the weight of saliency features $X$ are built on scale-wise information. Multi-scale features are obtained by stepwise down-sampling of $X$ for generating low resolution feature maps $ X^n\in R^ {\frac{M}{2^n} \times {\frac{N}{2^n}}\times C} ;(n = 1,2,3,\ldots,N)$, for $i\in 1,..., \frac{M}{2^n}\times\frac{N}{2^n}$.  
\setlength\abovedisplayskip{10pt}
\begin{equation}
\ell_i^n = p(L = i|X^n) = \frac{exp(W_i^n \times X_i^n)}{\sum^{{\frac{M}{2^n}}\times{\frac{N}{2^n}}}_{j=1} exp(W_j^n \times X_i^n)}
\end{equation} 
in which $W_i^n$ denotes the weights of the hidden layer, $j=1$ is a constant value, $L$ represents a random variable; $\ell$ is the attention map, where $\Sigma_i^{{\frac{M}{2^n}}\times {\frac{N}{2^n}}}\ell_i =1$. With these operations, DSF learns from attention maps of different regions at different scales. This is required for saliency segmentation as salient areas need higher weights. Moreover, stacked pooling is applied in pyramid units to improve the receptive field of different feature extraction layer, and feature representation can be enhanced by merging the features of different regions
\begin{equation}
\begin{aligned}
\Upsilon_j&=\underbrace{\frac{1}{N}\sum\limits_{n=1}^N \ell_j^n X_j,(j\in 1,\dots,(M\times N))}_\text{term (a)} \\[-3ex]
&=\underbrace{\frac{1}{N}\sum\limits_{n=1}^N (1+\ell_j^n)X_j, (j\in 1,\dots, (M\times N))}_\text{term (b)} 
\label{eq:short}
\end{aligned}
\end{equation}
in which $\Upsilon$ denotes the updated feature and $\Upsilon_j$ represents the $j^{th}$ feature map. As illustrated in Eq. \ref{eq:short} term (a), the model estimates the engaged parameters for the inputs by targeting the expectation over the image features in various areas. To reduce the number of null values in the feature maps and simplify the back-propagation we adopt identity mapping~\citep{he2016deep} as presented in Eq. \ref{eq:short} term (b). By adopting the residual connection we can preserve the input features. The outputs of the $K$ parallel dilated convolution are merged to generate $multi-dimensional$ feature map, Fig.~\ref{fig:2}{(b)} represents our approach. Our DSF module has $\frac{MN}{K}+\frac{{(nN)}^2}{K}$ parameters and the size of its receptive field is $[(n-1)2^{(k-1)}+1]^2$. In comparison with the ordinary convolution layers, factorization significantly moderates the number of parameters by a factor of $\frac{n^2MK}{M+{n}^2N}$. Thus, the DSF unit learns $\sim2.5\times$ fewer parameters with a receptive field of $15\times15$ than an ordinary convolution with a $3\times3$ receptive field, where $K=4$, and $N=M=128$.\\

{\bf Feature Pyramid Module}: $X$ represents the feature from the last convolution layers of the DSF units. Each $X^\ell$ is firstly downsampled into various scales. The network attention is calculated over three sequential operations: Batch Normalization (BN) is followed by $(1\times 1, 1)$ convolution and ReLU, with the $14\times 14$ attention maps. To resize the attention maps $(\ell^n)_n$ to its original size, upsampling is applied over all the scales to achieve an enhanced saliency representation $X^\ell$ via Eq.(\ref{eq:short}). To enhance the semantic of features in each level of the pyramid network, a weighting module is applied. Assume $F_n$ are the feature maps in $n^{th}$ pyramid level. If $H_n$ and $W_n$ represent the height and width of $F_n$, respectively, and $F_n^i$ denotes the $i^{th}$ channel feature. We compute the global distribution response $Z_n^i$ by using global average pooling on $F_n^i$ , as follows:
\begin{equation}
Z_n^i = \frac{1} {H_n \times W_n}\sum\limits_{p}^{H_n}\sum\limits_{q}^{W_n} F_n^i (f, q)
\end{equation} 
In order to compute the weight, two $1\times 1$ convolution layers are used to map the non-linear correlation among distribution responses $Z_n$ and acquire the significant feature vectors  $\hat Z_n = W_n^1 (\delta (W_n^2 Z_n))$,   
in which $W_n^1$  and $W_n^2$ are the weights of the first and the second convolution layers respectively and $\delta$ signifies the activation function. We use Sigmoid $\sigma$ to normalize $\hat Z_n$ as a weight vector $r_n=\sigma(\hat Z_n)$, and obtained the channel-wised feature weight $Fcr_n$, as follows:
\begin{equation}
Fcr_n = F_n r_n = [F_n^1 r_n^1, F_n^2 r_n^2,..., F_n^n r_n^n ]        
\end{equation}

Moreover, to improve the features, we define a function $F_n^{(f,q)}$ to unify the channel features at each point $(f, q)$ on $F_n$ \citep{zhou2021limited}. Firstly, the pointwise addition is performed on channel features to obtain the spatial vector $V_n^{(f, q)}$ while $W_n^3$ denotes the kernel weight of the convolution layer as: $V_n^{(f, q)} = W_n^3 F_n^{(f, q)}$. To generate the weight $t_n$ , we normalize the spatial vector $ V_n$ as $t_n =\sigma (V_n)$, therefore, the new spatial feature weight $Fsr_n$ is written as:

\begin{equation}
\begin{aligned}
Fsr_n = F_n t_n = [F_n^{(1, 1)} t_n^{(1, 1)}, F_n^{(1, 2)} t_n^{(1, 2)},...,\\
F_n^{(H_n, W_n)} t_n^{(H_n, W_n)} ] 
\end{aligned}       
\end{equation}

\begin{wrapfigure}{l}{0.5\textwidth}
\includegraphics[width=0.5\textwidth]{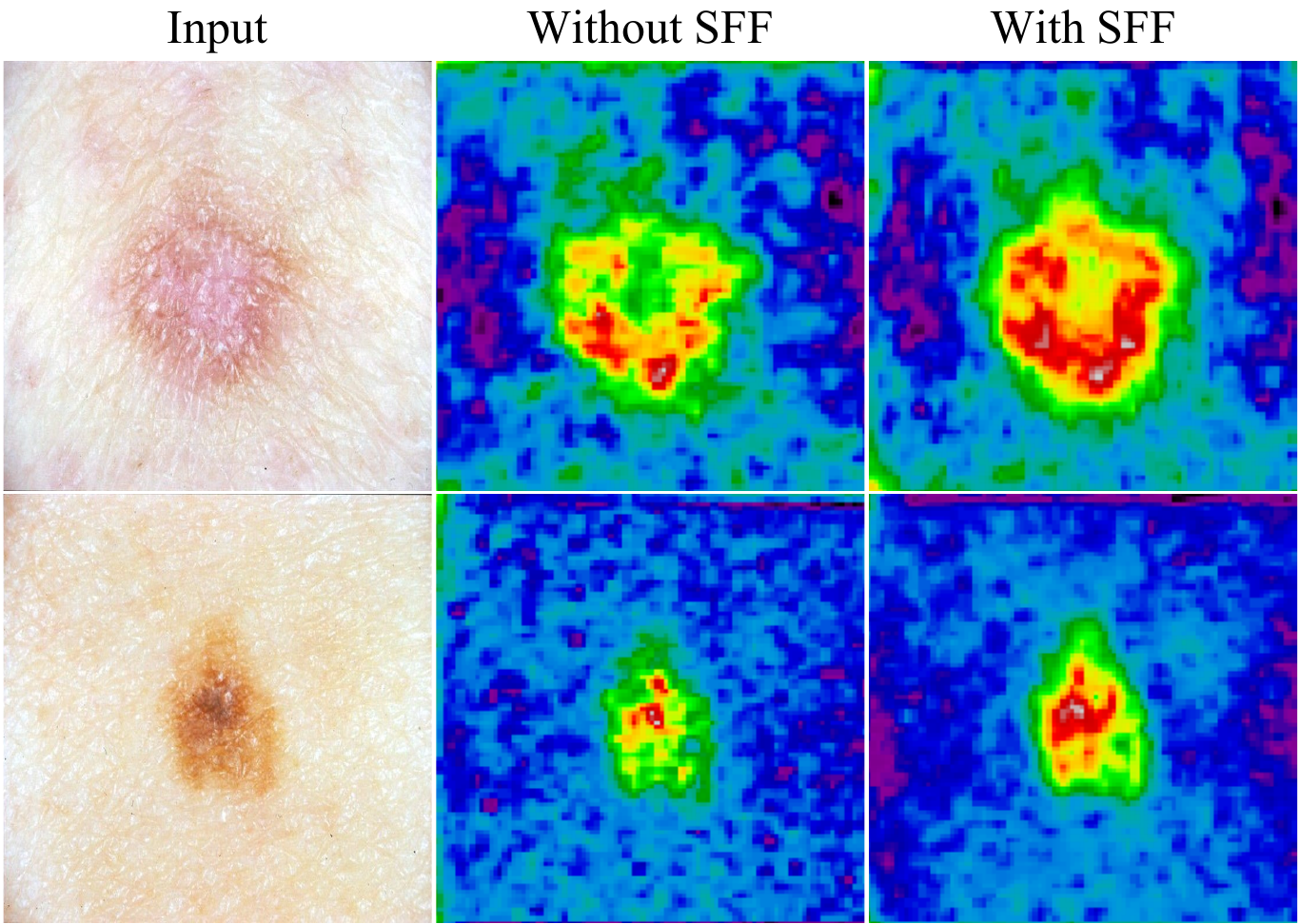}
\caption{Comparison of DSF module feature maps {\it with \& without} stepwise feature fusion (SFF). SFF rectifies the gridding artifact.}
\label{fig:3}
\end{wrapfigure}
\begin{figure*}
\centering
\includegraphics[width=0.99\textwidth]{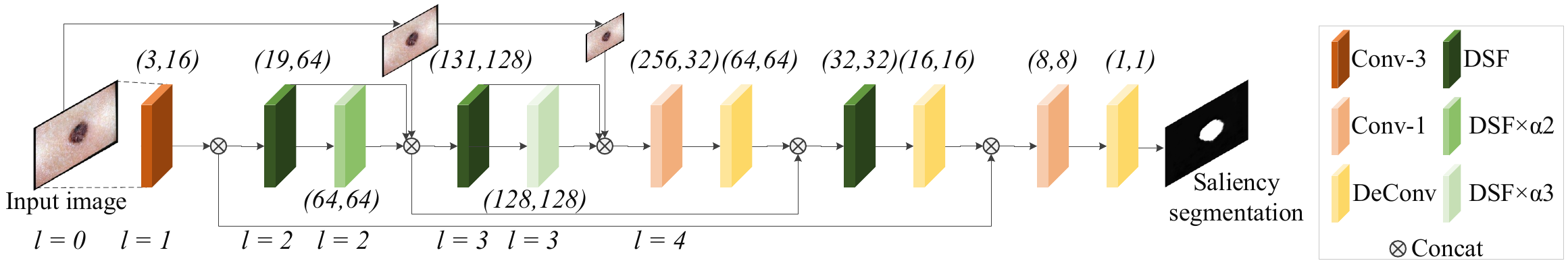}
\caption{The architecture of {\it DSF-Net}. The $Conv-3$ and {\it DSF} modules perform down-sampling and $DeConv$ module performs up-sampling actions. For each layer the {\it input channels} and {\it output channels} are shown.}
\label{fig:4}
\end{figure*}

Finally, we combine the channel-wised feature weight $Fcr_n$, and the spatial feature weight $Fsr_n$ to make an enhanced feature $Fr_n = Fcr_n + Fsr_n$ to capture more important information.\\
{\bf Stepwise Feature Fusion (SFF)}: By combining the results of the dilated convolutions, the proposed DSF unit provides a large receptive field. However, it generates some undesirable artifacts, as shown in Fig.~\ref{fig:3}. To get rid of these artifacts, we propose to merge the obtained feature maps from different kernel scales and dilation rates before concatenating them, as shown in Fig.~\ref{fig:2} that helps to pay more attention to the global information and further refines the focus area of the model. This effective and simple solution will not increase the network complexity. The network only uses an element-wise addition  to combine the input and output feature maps~\citep{he2016deep}.\\
{\bf DSF-Net architecture}: Except the first layer where standard convolution is used, in the rest of proposed network, DSF modules are adopted for learning and down-sampling tasks. All the convolution layers and the DSF units are followed by {\it BN}~\citep{ioffe2015batch} and {\it ReLU}~\citep{he2015delving}. Then the output features are processed by a $1\times 1$ convolution and sigmoid for generating the saliency map. To upscale the saliency map into its original size we use bilinear interpolation. Fig.~\ref{fig:4} in detail illustrates the architecture of DSF-Net. The saliency segmentation outputs have the same size as that of the input image and a light weight decoder is used. A hyperparameter $\alpha$ controls the depth of the network to create a deep efficient network while preserving the network topology. Generally CNNs demanding additional memory at higher spatial levels due to the high dimensions of feature maps. However, neither the convolution layer nor the DSF units are repeated at these stages.\\

{\bf Loss Function}: To improve saliency prediction and training, a fused loss function is proposed  on the basis of various evaluation metrics. Let $G\in [0, 1]_{r\times q}$ and $S\in [0, 1]_{r\times q}$ be the input saliency map and the estimated saliency respectively, in which $r\times q$ denotes the resolution of the saliency map, and the final loss $\mathcal L$ computed as: 
\begin{equation}
\mathcal L(G,S) = \mathcal L_{Cross entropy} (G, S) + \mathcal L_{MAE} (G, S)
\end{equation}
in which $\mathcal L_{cross entropy}$ and $\mathcal L_{MAE}$ specify {\em cross entropy} and {\em MAE loss} respectively. $\mathcal L_{MAE}$ is based on the {\em MAE} metric, which is widely used in the identification of salient objects. $\mathcal L_{MAE}$ calculates the variance between the predicted saliency map $S$ and the corresponding ground truth $G$.

\begin{table*}[h]
\centering
\scriptsize{
\caption{Quantitative evaluation with $F_{score}$ and RPI (higher is better), MAE (lower is better) over four skin lesions datasets. For each column, the best result is highlighted.}
\begin{tabular}{|p{.1\textwidth}|| p{.037\textwidth}p{.037\textwidth} p{.04\textwidth}| p{.037\textwidth}p{.037\textwidth} p{.04\textwidth}| p{.037\textwidth}p{.037\textwidth} p{.04\textwidth} | p{.037\textwidth}p{.037\textwidth} p{.04\textwidth}|}
\hline
\multirow{2}{*}{Methods} & \multicolumn{3}{c|}{ISBI-2016} & \multicolumn{3}{c|}{ISBI-2017} & \multicolumn{3}{c|}{PH2} & \multicolumn{3}{c|}{ISIC-2018} \\ \cline{2-13} 
& $F_{score}$  & MAE  & PRI   &$F_{score}$   & MAE   & PRI   & $F_{score}$    & MAE   & PRI & $F_{score}$    & MAE   & PRI  \\ 
\hline
mDRFI  &0.729&	0.056&	0.732&	{0.714}&	0.058	&0.739&	0.691&	0.052&	0.708&0.721&	0.051&	0.737 \\   [- 0.1ex]
eVida  &0.742&	0.059&	0.757&	0.726&	0.059&	0.733&	0.705&	0.054&	0.712& 0.737&	0.053&	0.748 \\     [- 0.1ex]
YGC  &0.816&	0.051&	0.825&	0.803&	0.054&	0.818&	0.759&	0.043&	0.766 &0.821&	0.052&	0.819 \\    [- 0.1ex]
DDNet  &	0.878&{0.043}&	0.889&	0.863&\bf{0.043}&	0.874& {0.843}&{0.034}& 0.851&	0.869&{0.045}&	0.876 \\  [- 0.1ex]
PAGENet  &{0.786}&	0.048&	{0.795}&	0.769&	0.048&	0.772&	0.752&\bf{0.032}&	0.760& {0.775}&	0.047&	{0.781} \\  [- 0.1ex]
SRM &0.731&{0.044}&	0.739&	0.698&	0.047& {0.706}&	0.679&	0.039& {0.685} &0.713&{0.046}&	0.722 \\   [- 0.1ex]
FCN-AlexNet  &0.904&{0.041}&	0.916&	0.889&	0.044& {0.895}&	0.872&	0.034& {0.879}&0.895&{0.043}&	\bf{0.903} \\   [- 0.1ex]
{\bf DSF w/o SFF} & {0.856}& {0.053}& {0.862}& {0.838}& {0.049}& {0.845}& {0.819}& {0.037} &  {0.827}& {0.842}& {0.054}& {0.851}\\    [- 0.1ex]
{\bf DSF-Net} & \bf{0.908}&\bf{0.040}&\bf{0.919}&\bf{0.894}&{0.046}&\bf{0.901}& \bf{0.876}&{0.033} & \bf{0.882} & \bf{0.897}&\bf{0.042}&{0.901}\\  [- 0.1ex]
\hline
\end{tabular}}
\label{tab:1}
\end{table*} 
\begin{table*}[h]
\centering
\scriptsize{
\caption{Quantitative evaluation with VOI, GCE and BDE (for all metrics lower is better) over four skin lesions datasets. For each column, the best result is highlighted.}
\begin{tabular}{|p{.1\textwidth}|| p{.037\textwidth}p{.037\textwidth} p{.04\textwidth}| p{.037\textwidth}p{.037\textwidth} p{.04\textwidth}| p{.037\textwidth}p{.037\textwidth} p{.04\textwidth} | p{.037\textwidth}p{.037\textwidth} p{.04\textwidth}|}
\hline
\multirow{2}{*}{Methods} & \multicolumn{3}{c|}{ISBI-2016} & \multicolumn{3}{c|}{ISBI-2017} & \multicolumn{3}{c|}{PH2} & \multicolumn{3}{c|}{ISIC-2018} \\ \cline{2-13} 
& VOI  & GCE   &BDE   & VOI   & GCE   &BDE & VOI  & GCE   &BDE   & VOI   & GCE   &BDE    \\ 
\hline
mDRFI&	1.593&	0.167&	16.07& 1.684&	0.173&	17.26&	1.698&	0.194&	17.41&	1.664&	0.169&	16.57 \\    [- 0.1ex]
eVida& 1.579&	0.164&	14.32 & 1.673&	0.169&	15.52	& 1.683&	0.181&	15.69& 1.662&	0.166&	15.46 \\   [- 0.1ex]
YGC &	1.574&	0.155&	12.37 & 1.663&	0.166&	13.06& 1.678&	0.186&	13.22	& 1.582&	0.159&	12.48\\   [- 0.1ex]
DDNet &	1.567&	0.154&{9.46} &1.654&\bf{0.148}&	10.73	&1.672&{0.166}&	10.88&	1.574&	0.160&\bf{9.50}	 \\   [- 0.1ex]
PAGENet &		1.569&	0.155&	10.86& 1.667&	0.162&	{11.41}& 1.676&	0.179&	{11.63}&		1.575&	0.156&	10.92 \\   [- 0.1ex]
SRM &{1.565}&{0.149}&	9.51& {1.654}&	0.159&	\bf{10.09}& {1.669}&	\bf{0.161}&	10.39&{1.569}&{0.153}&	9.59 \\  [- 0.1ex]
FCN-AlexNet &	1.558&	\bf{0.148}&	9.43& {1.652}&	0.156&	10.38& 1.667&	0.168&	10.59&	1.573&	\bf {0.151}&	9.52 \\  [- 0.1ex]
{\bf DSF-Net} &\bf{1.556}&{0.149}&\bf{9.41} & \bf{1.651}&{0.152}&{10.11} & \bf{1.663}&{0.163}&\bf{10.34}&\bf{1.562}&{0.152}&{9.51}\\ [-0.1ex]
\hline
\end{tabular}}
\label{tab:2}
\end{table*} 

\section{Experiments and Protocols}
To evaluate the proposed DSF-Net, four datasets have been used ISBI-2016, ISBI-2017, PH2, and ISIC-2018. The ISBI-2016 contains 1,279 dermoscopic images and the ISBI-2017 \citep{gutman2016skin} consists of 23,906 images. PH2 dataset~\citep{mendoncca2013ph}, contains 200 dermoscopic images and ISIC-2018 dataset \citep{codella2018skin} contains 2594 dermoscopic images. For a comprehensive evaluation, we use several metrics: {\it Precision-Recall} (PR) curves, {\it F-measure}, {\it Mean Absolute Error} (MAE),  {\it Variation of Probabilistic Rand Index} (PRI), {\it Information metric} (VOI), {\it Global Consistency Error} (GCE), and {\it Boundary Displacement Error} (BDE) \citep{shamsolmoali2020road}. A segmentation result is satisfactory based on these parameters when the comparison with ground truth gives high values for the {\it F-measure} and PRI, but low values for the other metrics.
Pytorch is used for all the implementations. Based on the training protocol shown in \citep{unver2019skin}, we used ISBI-2017 for training. The stochastic gradient descent optimizer is used to train our proposed model with a batch size of 16. The learning rate is set to 0.001 and it is decreased by a factor of 0.01 after 100 and 200 epochs. Moreover, we have set the weight decay factor to 0.0005 and the momentum to 0.9. We trained the model for 300 epochs which takes around 4 hours using an Nvidia GTX 1080 GPU. DSF-Net processes an input image in $0.45s$ which is faster than the current salient segmentation based CNN approaches.

\section{Results and Analysis}
We evaluate the performance of DSF-Net against the other recent approaches such as: mDRFI~\citep{jahanifar2018supervised}, eVida~\citep{garcia2019segmentation}, YGC~\citep{unver2019skin}, DDNet~\citep{li2018dense}, PAGENet~\citep{wang2019salient}, SRM~\citep{wang2017stagewise}, and FCN-AlexNet~\citep{kaymak2020skin}. 

\begin{figure*}[h!]
\centering
\includegraphics[width=0.9\textwidth]{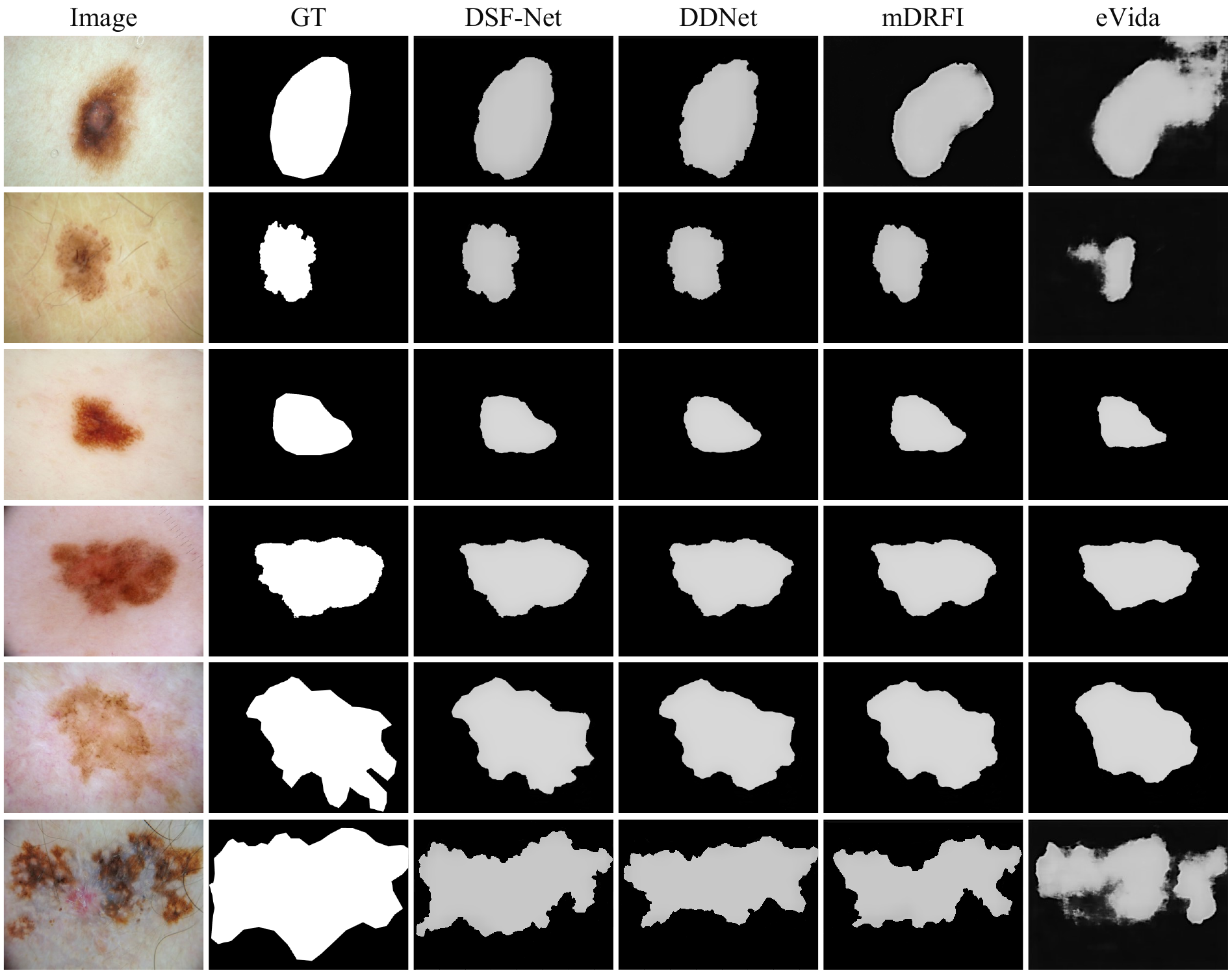}
\caption{Segmentation and visualize results for different methods on some representative examples.}
\label{fig:6}
\end{figure*} 
In Table~\ref{tab:1} we report the test results of DSF-Net and the other approaches on all four datasets in terms of F-measure, MAE and PRI scores. DSF-Net considerably improved F-measure compare to the other approaches. Moreover, in Table~\ref{tab:2} we report the test results of DSF-Net and the other approaches in terms of VOI, GCE, and BDE scores.
The results validates the superior performance of DSF-Net in different scenes. In Fig.~\ref{fig:6} we visualize the segmentation results of different models. From this results we observe DSF-Net has better performance in a variety of complex situations, for example, large lesions, while there is low contrast between lesions and the surrounding skin, and multiple disconnected lesions. Furthermore, DSF-Net captures boundaries properly because of using salient edge detection. Here, we evaluate the contribution of each module to the overall performance of the proposed DSF-Net. The experiments conducted on the ISBI-2016 and ISBI-2017. 
\begin{figure}
\centering
\includegraphics[width=0.71\textwidth]{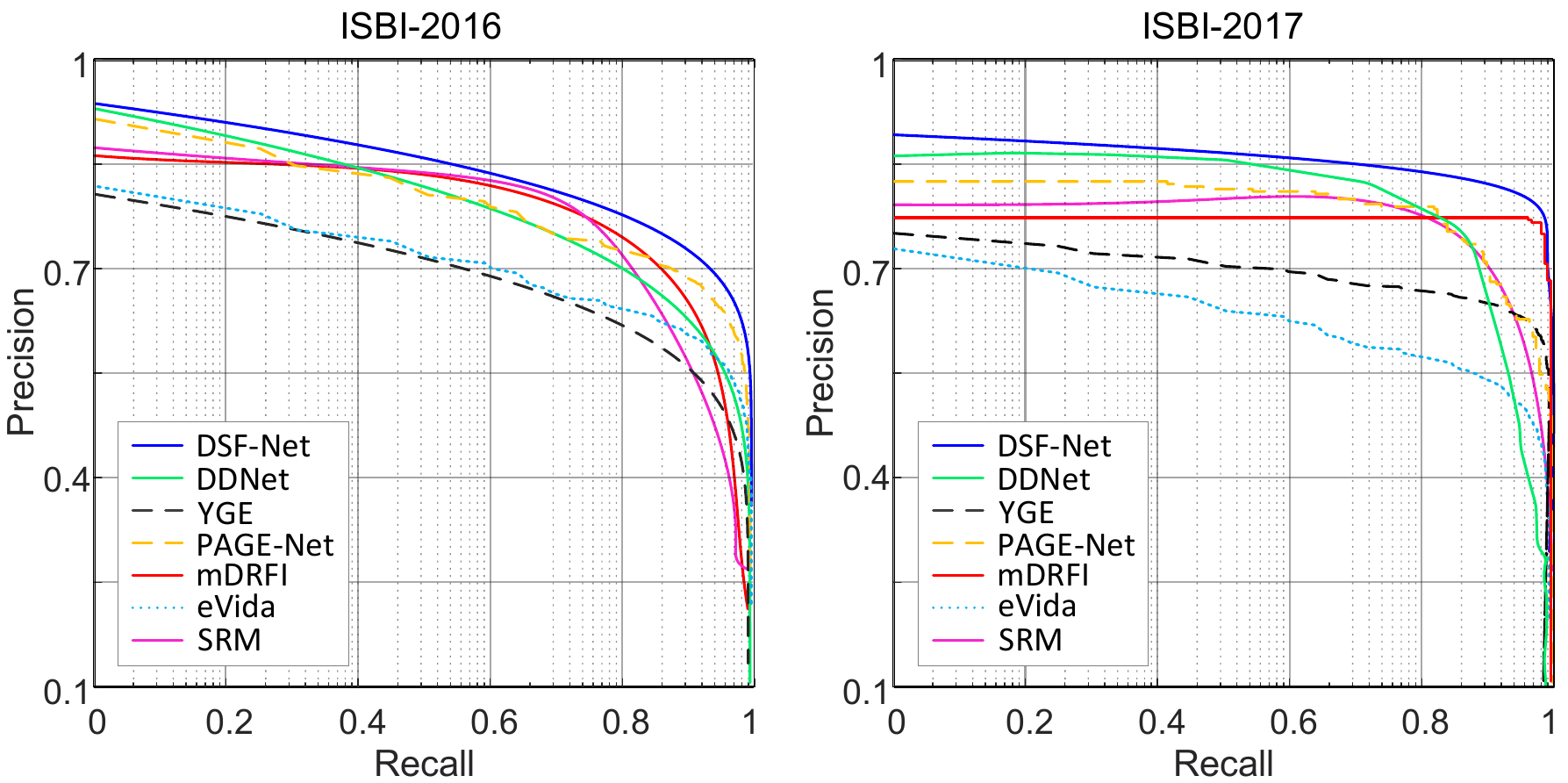}
\caption{ Quantitative comparison of six models using PR curve on ISBI-2016 and ISBI-2017 datasets.}
\label{fig:9}
\end{figure}
\begin{figure}
\centering
\includegraphics[width=0.71\textwidth]{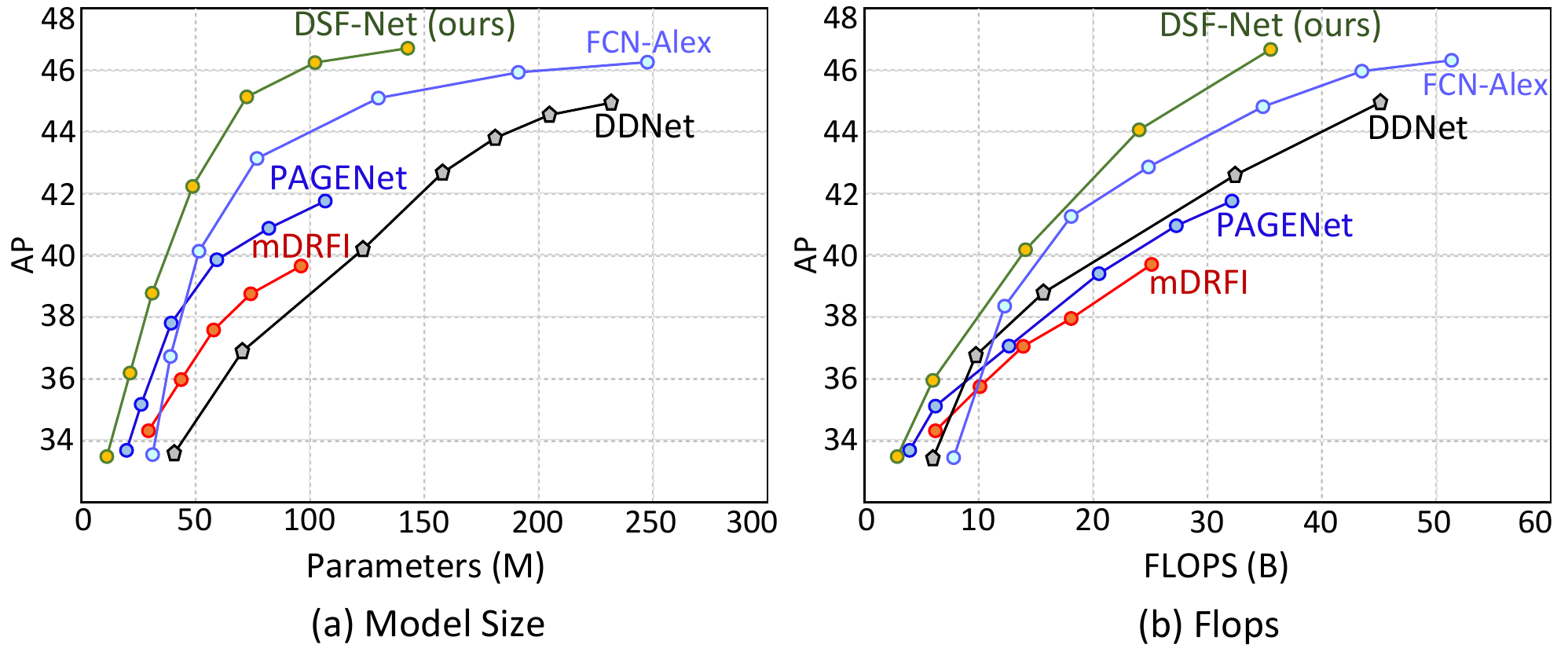}
\caption{ Efficiency comparison of state-of-the-art models.}
\label{fig:8}
\end{figure}

\indent {\bf Down-sampling and Residual learning}: Replacing the standard convolution with the strided DSF and skip-connections in the proposed DSF-Net improved accuracy by 3\% with 26\% parameter reduced. This verifies the effectiveness of the selected learning scheme. The PR curves of all approaches are shown in Fig.~\ref{fig:9} on two datasets. We observe that DSF-Net achieved a higher ROC score (4.2\% better) comparing to that obtained by other models in the testing set. The fact that DSF-Net not only learns highly abstract salient features of skin lesions with a lower number of parameters but also results in improving semantic segmentation and can be use for lesion segmentation. \\

\indent {\bf Effect of different convolutions in the DSF modules}: The DSF module use instant convolutions to rotate and reduce the dimensional of feature maps whilst using dilated pyramid convolutions to transform the feature maps. To show the effect of these two components, the following experiments are performed. Instant convolutions: standard convolutions with rotated matrix are substituted with dilated pyramid convolutions in the DSF module; the resultant network is more efficient with a less number of parameters and improves the saliency segmentation by $1.5\%$. Fig.~\ref{fig:8} shows the performance of DSF-Net in comparison with the other state-of-the-art models. DSF modules outperformed mDRFI and PAGENet by $1.14\%$ and $1.08\%$, respectively, while learning the same number of parameters with lower network sizes and higher speeds. In addition to the parameter size and FLOPs, we have also evaluated the latency of the algorithms. Each model runs $5$ times with the batch size $1$, the standard deviation and mean are reported. Fig.~\ref{fig:8} reveals the performance on the model size, flops and GPU latency. In comparison with the other state-of-the-art models, DSF-Net can make 1.6$\times$ faster when using GPU. Overall, DSF-Net shows satisfactory segmentation performance with better boundary adherence and less displacement errors in these experiments.
\section{Conclusion}
We have introduced a novel saliency segmentation model, DSF-Net, based on an efficient dilated pyramid network. The proposed model improves feature extraction and saliency representations with stepwise dilated convolution. In addition, instant convolution enables more efficient training and better performance. Extensive experimental evaluations on three benchmark datasets demonstrate that our algorithm significantly improved segmentation performance. In addition, the proposed DSF-Net architecture has better efficiency when using GPUs for acceleration.

\bibliography{iclr2019_conference}
\bibliographystyle{iclr2019_conference}

\end{document}